\documentclass[sigconf]{acmart}
\usepackage{algorithm}
\usepackage{algorithmic}
\usepackage{graphicx}

\begin{document}

\title{JTreeformer: Graph-Transformer via Latent-Diffusion Model for
Molecular Generation}
  \author{Ji Shi}
 \affiliation{
   \institution{Harbin Institute of Technology, Shenzhen}
   \city{Shenzhen}
   \country{China}
   }
 \email{sj1402904919@gmail.com}

   \author{Chengxun Xie}
   \authornote{Equal contribution.}
 \affiliation{
   \institution{Harbin Institute of Technology, Shenzhen}
   \city{Shenzhen}
   \country{China}
   }
 \email{1085959318@qq.com}

 \author{Zhonghao Li}
 \affiliation{
   \institution{Harbin Institute of Technology, Shenzhen}
   \city{Shenzhen}
   \country{China}
   }
 \email{lizhonghao@stu.hit.edu.cn}

   \author{Xinming Zhang}
 \affiliation{
   \institution{Harbin Institute of Technology, Shenzhen}
   \city{Shenzhen}
   \country{China}
   }
 \email{xinmingxueshu@hit.edu.cn}
    \author{Miao Zhang}
    \authornote{Corresponding author.}
 \affiliation{
   \institution{Harbin Institute of Technology, Shenzhen}
   \city{Shenzhen}
   \country{China}
   }
 \email{miaozhang1991@gmail.com}



\begin{abstract}
The discovery of new molecules based on the original chemical molecule distributions is of great importance in medicine. The graph transformer, with its advantages of high performance and scalability compared to traditional graph networks, has been widely explored in recent research for applications of graph structures. However, current transformer-based graph decoders struggle to effectively utilize graph information, which limits their capacity to leverage only sequences of nodes rather than the complex topological structures of molecule graphs. This paper focuses on building a graph transformer-based framework for molecular generation, which we call \textbf{JTreeformer} as it transforms graph generation into junction tree generation. It combines GCN parallel with multi-head attention as the encoder. It integrates a directed acyclic GCN into a graph-based Transformer to serve as a decoder, which can iteratively synthesize the entire molecule by leveraging information from the partially constructed molecular structure at each step. In addition, a diffusion model is inserted in the latent space generated by the encoder, to enhance the efficiency and effectiveness of sampling further. The empirical results demonstrate that our novel framework outperforms existing molecule generation methods, thus offering a promising tool to advance drug discovery.\footnote{ \url{https://anonymous.4open.science/r/JTreeformer-C74C}}
\end{abstract}






\maketitle

\section{Introduction}
\label{sec:intro}

Identifying novel molecules poses a significant challenge due to the vast chemical spaces involved, necessitating efficient search methods among theoretically feasible candidates. Instead of searching a tremendous space for exhaustive enumeration, deep generative models offer an alternative and promising solution by random sampling and exploring the properties of molecules.
Graph generative models introduce the graph representation method for molecules instead of the SMILES notation \cite{r2}, with nodes representing atoms and edges denoting chemical bonds \cite{r5, r6, r7}. Further research \cite{r8,r9} indicates that fragment-based representation, which treats chemical fragments as nodes of the graph, can address the generation of invalid molecules due to the diverse nature of chemical bonds. Traditional chemical molecular graph data are processed and encoded using Graph Neural Network based models (GNNs). Although performing well in small molecules, GNNs fully encapsulate the complex higher-order relationships and global structural nuances inherent in molecular graphs \cite{r0}. Variational Autoencoders (VAEs) based molecule generation also rely on GNNs, which are also constrained by the Weisfeiler-Lehmann isomorphism hierarchy and suffer from over-smoothing and over-squashing issues\cite{r27}, necessitating the adoption of a novel architecture for VAE. 

Graph transformer, due to self-attention mechanisms for a more dynamic and global understanding of graph structure\cite{r29}, has ignited fervent research activity in specific tasks of molecule classification or identification. Nevertheless, research of the decode method for transformers in generating molecules is still an open question, due to the lack of effective decoder methods tailored to attention mechanisms and graph information. Mitton et al. \cite{r32} use a transformer decoder to reconstruct the edges of the graph, demonstrating the potential of the transformer as a VAE graph VAE. Despite showing promising results, their model employs separate networks for node and adjacency matrix reconstruction, relying on GraphSAGE \cite{r33} and DIFFPOOL \cite{r34} for node reconstruction, which adds complexity to the model and hardly fully exploit the transformer's capabilities.

Although direct random sampling in the latent space is possible, we empirically find that latent diffusion models (LDMs) improve the quality of generated molecular samples. LDMs capitalize on a structured noise-reduction process that iteratively refines the sampled data, leading to more accurate and stable molecular configurations. This approach in latent space learning not only improves sample fidelity but also provides a more controllable generation pathway compared to conventional random sampling methods.

\begin{figure*}[ht]
  \centering
  \includegraphics[width=\textwidth]{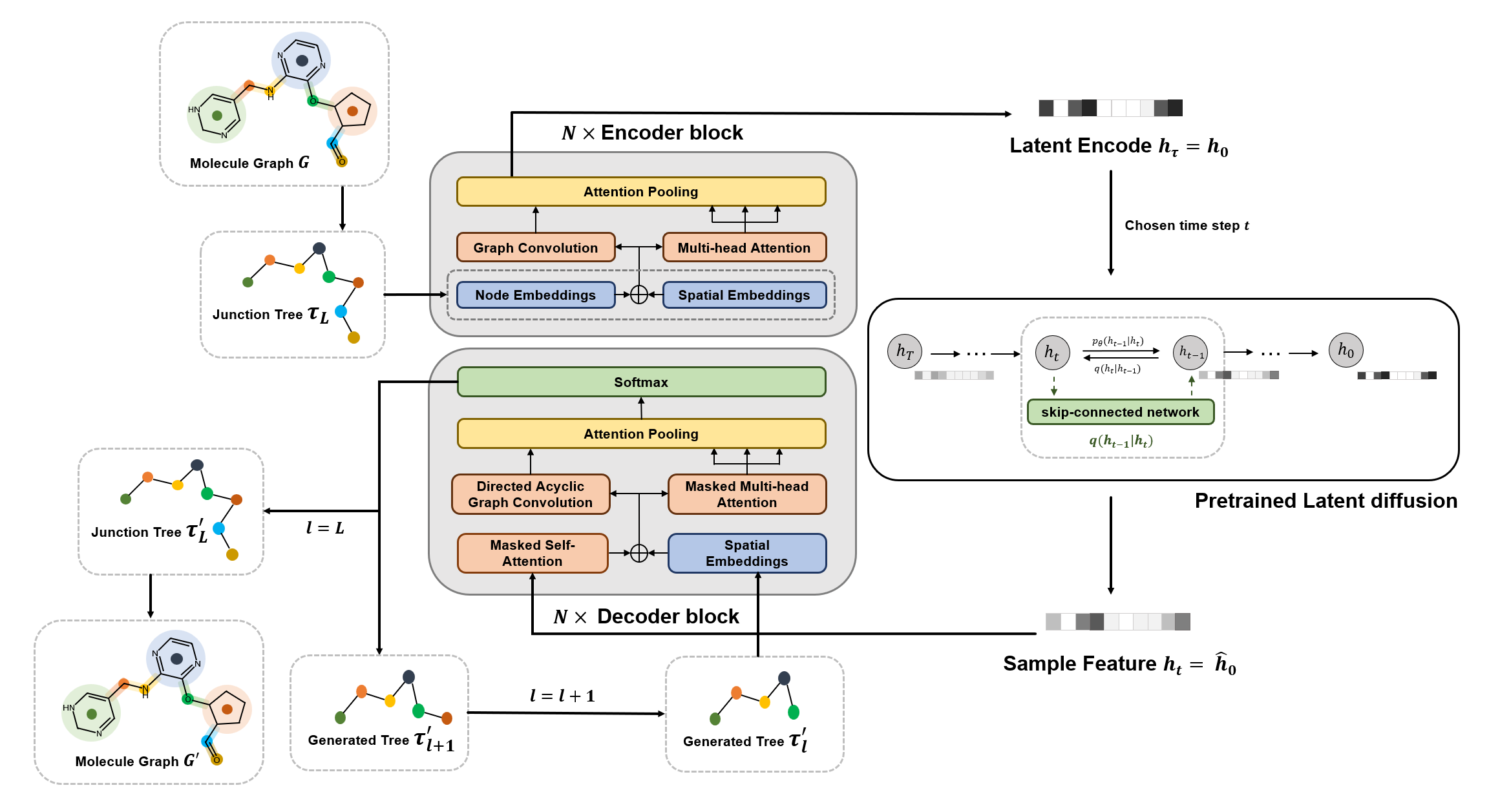}
  \caption{Architecture of JTreeformer,which involves an encoder-decoder structure with attention mechanisms and graph convolutions (DAGCN and mask-attention for decoder). In the latent space of the molecule generated by the encoder, the latent diffusion process is utilized to sample and generate molecular structures iteratively.}
  \Description{This image illustrates the architecture of the JTreeformer, a machine learning model designed for molecular structure generation. The model consists of an encoder-decoder structure with attention mechanisms and graph convolutions (DACGN and mask-attention for the decoder). The encoder block processes the input molecule graph (G) and junction tree (T_L), generating node embeddings and spatial embeddings. The decoder block samples from the latent space of the molecule and generates a new molecule graph (G') and junction tree (T_{l+1}). The latent diffusion process is used to iteratively sample and generate molecular structures.}
  \label{fig:molgenmodel}
\end{figure*}

In this paper, we present a novel framework for encoder-decoder structure called \textbf{JTreeformer}, a {G}raph-{T}ransformer via {L}atent-{D}iffusion {M}odel for Molecule Generation, depicted in Fig.\ref{fig:molgenmodel}, which gives an affirmative answer to the utilization of transformer mechanism in molecule generation and achieves state-of-the-art performance on a wide range of molecule generation tasks. Inspired by JT-VAE\cite{r8}, we transform graph generation into sequence generation to adapt the encoder by using junction trees in BFS order, which has only 2N-1 predictions required for reconstructing a graph with N nodes. We devise a hybrid network as the encoder which integrates both Graph Convolutional Networks (GCNs) and Transformers to learn a robust representation of molecule structure. In addition, we propose the Directed Acyclic Graph Convolution Network (DAGCN) paralleled with masked multi-head attention for the architecture of the decoder, which can iteratively synthesize the entire molecule by leveraging information from the partially constructed molecular structure at each step. Classic generative models, such as auto-regressive \cite{r16, r17}, VAE \cite{r7, r8, r12}, and GAN \cite{r13, r14}, share a common limitation: they fail to capture the original distribution of molecules, hindering molecular optimization tasks. On the contrary, diffusion models \cite{r19} have demonstrated a robust capacity for learning the original data distribution for image synthesis. Ignited by the success of diffusion model in image generation, we trained a diffusion model using Skip-Connected-Net as the noise prediction network in the latent space of JTreeformer and showed its proficiency in generating high-quality molecules.

In summary, we devised a novel encoder-decoder structure with attention mechanisms and graph convolutions for molecular generation, called \textbf{JTreeformer}, and main contributions are as follows:

\begin{itemize}
  \item We combine the GCN with Graph transformer for encoder blocks in our JTreeformer, which facilitates the integration of local information aggregation from GCNs and global information summarization from Graph Transformers, thereby generating more affirmative latent space of molecules.
  \item In our JTreeformer, we devise a Directed Acyclic Graph Convolutional Network (DAGCN) with a masked Graph transformer for decoder blocks for adapting the latent space acquired by the encoder. DAGCN is able to predict subsequent nodes in an incomplete graph, effectively remedying the traditional GCN's inability to forecast node sequences.
  \item We introduce the latent diffusion model to the domain of molecule generation for controlling the process, which offers the advantage of generating high-quality samples through a controlled and gradual reverse diffusion process, effectively capturing the intricate distribution of vast chemical space.
  \item Experimental results show the proposed method outperforms the fully supervised baselines. Visualizing the result of learned latent embedding confirms that the proposed method can reveal the pattern of vast chemical space.
\end{itemize}

\section{Related Work}
\textbf{Junction tree based VAE.} In \cite{r8}, Jin et al. propose JT-VAE that leverages junction trees for representing a fragment-based graph-represented molecular generation. It extracts subgraphs from molecular graphs and converts them into a junction tree, with subgraphs serving as nodes. The model sequentially generates the entire junction tree on a subgraph-by-subgraph basis, subsequently reassembling these trees into molecules. These subgraphs encompass singletons, bonds consisting of two atoms, and rings. JT-VAE has demonstrated the capability to generate 100\% chemically valid molecules. Nevertheless, JT-VAE employs RNN for message passing, leading to challenges with long-range forgetting and the inability to parallelize training. Additionally, reassembling the junction tree into molecules generates numerous isomers, which requires an additional neural network to predict the most probable ones \cite{r8} and incurs significant training costs. 

\noindent\textbf{Language model for graph generation.} The autoregressive generation of graphs has sparked interest in applying language models to graph generation, necessitating the sequentialization of graphs as the input and output of the model. A seminal work is GraphRNN by You et al. \cite{r16}, which reformulates graph generation as two sequence-to-sequence tasks: (1) predicting the next node based on all previous nodes and (2) predicting the edges connecting the next node with all preceding nodes. This approach requires $\frac{n(n+1)}{2}$ predictions for a graph with n nodes. To alleviate this complexity, the authors use Breadth First Search (BFS) to prioritize node ranking. This ensures that each node's parent is proximate, limiting edge prediction to a fixed-length window, thus reducing predictions to O(n). However, when connected nodes reside outside this window, GraphRNN's ability to accurately predict adjacency relationships is compromised.

\noindent\textbf{Graph transformer.} Recently, the transformer has become the most powerful tool in building language models. The self-attention mechanism of the transformer \cite{r30} can be interpreted as an MPNN operating on a fully connected graph, thereby bypassing the limitations of traditional GNNs and emerging as a viable alternative. Ying et al. \cite{r29} introduced Graphormer, a pure transformer model that surpassed GNN-based methods in multiple graph prediction tasks, highlighting the transformer's proficiency in handling graphs. The integration of transformers and GNNs has also garnered significant attention, with Min et al. \cite{r31} summarizing these efforts. They identified three primary categories: GNNs as Auxiliary Modules (GA), Improved Positional Embedding from Graphs (PE) improved through graphs and Improved Attention Matrices from Graphs (AT), and explored the performance gains achieved by each approach. While most studies utilize the transformer encoder to extract graph information, Mitton et al. \cite{r32} proposed the use of the transformer decoder to reconstruct graph edges, demonstrating the transformer's potential as a graph VAE. However, their model employed separate networks for node and adjacency matrix reconstruction, relying on GraphSAGE \cite{r33} and DIFFPOOL \cite{r34} for node reconstruction. This approach added complexity to the model's architecture and did not fully exploit the transformer's capabilities.

\noindent\textbf{Diffusion Model (DM).} DDPM \cite{r19} introduced a method where Gaussian noise is continuously added to the original image until it transforms into pure noise. The theory of stochastic differential equations (SDEs) suggests that this process is reversible, enabling the reconstruction of the original image from pure noise by training a model to predict the noise added at each step. Rombach et al. \cite{r20} proposed a method that utilizes VAE to initially map images into a low-dimensional latent space, followed by training a diffusion model in latent space. This approach enhances generation granularity while reducing training costs. For a conditional generation, Dhariwal et al. \cite{r21} introduced classifier guidance, which necessitates the training of an additional classifier alongside the unconditional DM. Song et al. \cite{r24} proposed DDIM, a novel generation method based on extrapolation, which enables denoising to be performed only on a subsequence of the complete denoising process, significantly accelerating the generation.

\section{JTreeformer}
In this section, we describe our JTreeformer, a framework of Variational Auto-Encoder (VAE) for molecule generation in the order of encoder, latent space learning, and decoder.

More specifically, Graph Convolutional Network (GCN) paralleled with the attention mechanism is designed in the encoder for empirically leveraging both local and global information from graphs. Then, the Directed Acyclic Graph Convolution Network (DAGCN) is devised for generation by gathering information of previously generated nodes in the adapted decoder, paralleled with masked multi-head attention. After encoder-decoder training, we further insert a skip-connected diffusion model \cite{r36, r37} as the noise prediction network for latent space learning, efficiently sampling in the virtual molecule space via the controlled process. To efficiently assemble junction trees into molecules, we employ Monte Carlo Tree Search (MCTS) \cite{r38} to eliminate the isomer problem. Algorithm \ref{alg::train} and \ref{alg::test} display the pertaining and decoding of JTreeformer in molecule generation.

\subsection{Modeling molecules as sequences}
\label{sec:model}
In JTreeformer, we first convert the molecule graphs to learnable sequences as junction trees, which helps to keep the validity of molecule in chemistry\cite{r8}. Then we devise a new method for capturing learnable sequences on downstream tasks from junction trees.

An undirected graph $G(V,E)$ is defined by its node set $V=(v_1,v_2,\dots,v_n)$ and its edge set $E = \{(v_i,v_j) | v_i,v_j \in V\}$. Junction $J$ is defined as a sub-graph of an undirected graph, and junction tree $JT(J,E)$ is a tree structure of junctions. We then define a mapping $f_{J \to G}$ from graphs to junction trees. For a graph $G \sim p(G)$, we have:
\begin{equation}
\begin{split}
\label{eq1}
    JT(J,E') = f_{G \to J}(G(V,E)).
\end{split}
\end{equation}

We define a permutation $\pi$ to traverse the junction tree to convert it into a learnable sequence, where mapping $f_{J \to S}$ is from junction trees to sequences:
\begin{equation}
\begin{split}
\label{eq2}
    S_n^\pi = f_{J \to S}(JT | \pi) = (s_1^\pi,s_2^\pi,\dots,s_n^\pi),
\end{split}
\end{equation}
where $s_{i+1}^\pi = (J_{i+1},P_{i}^\pi) \in \mathbf{R}^{V_J \times P}$. $J_{i} \in V_J$ denotes the $i^{th}$ specific junction in the junction vocabulary $V_J$, and $P_{\pi, i}$ denotes the kind of positional relationship between the current node and its father node, which is determined by the previous node in the permutation $\pi$. 

\begin{figure}[t]
  \centering
  \includegraphics[width=3in]{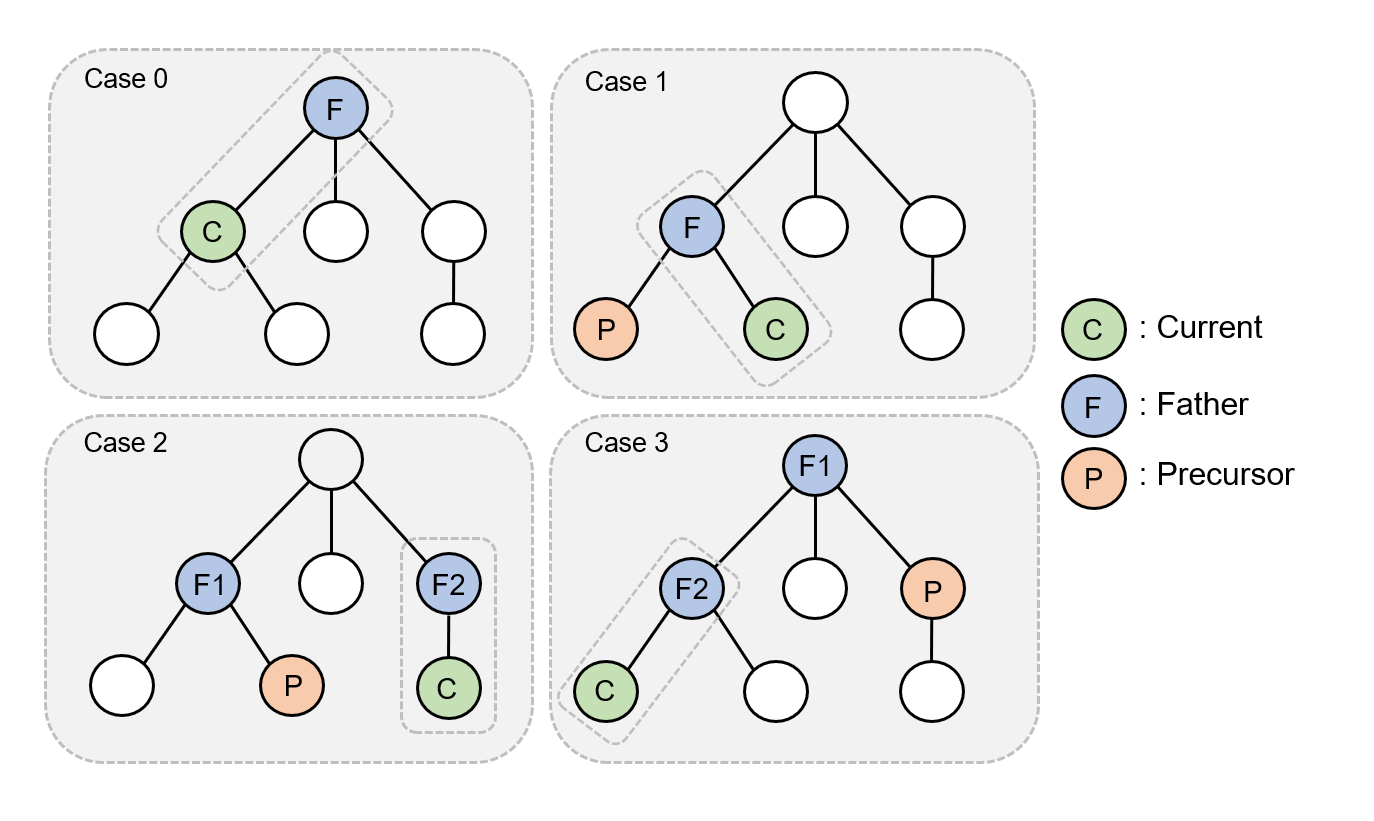}
  \caption{An example of prediction in junction node position.  F1 denotes the father of the precursor, while F2(F) denotes the father of the current node.
  Case 0 indicates that the father and precursor of the current node are the same node. Case 1 indicates that precursor and current share the same father node, while Case 2 and 3 are two examples to illustrate the special relationship between current and its father F2. }
  \label{pic::BFS}
  \Description{This figure presents four distinct tree-structured scenarios (Case 0 to Case 3), each enclosed in a dashed outline to indicate a separate example. Within each tree, circles represent nodes. Three special nodes are highlighted in color: C (green) for “Current,” F (blue) for “Father,” and P (orange) for “Precursor.” The remaining, uncolored circles represent generic nodes without special roles.

    Case 0: A single Father node (F) is directly connected to the Current node (C), which then links to other nodes.
    Case 1: The Precursor node (P) and the Current node (C) are both connected to the Father node (F), forming a small substructure distinct from the other nodes in the tree.
    Case 2: Two Father nodes (F1 and F2) appear, with the Precursor (P) branching from F1 and the Current node (C) branching from F2.
    Case 3: Similar to Case 2, there are two Father nodes (F1 and F2). However, in this arrangement, the Current node (C) extends from F2 and the Precursor (P) directly attaches to the main structure under F1.
    Each dashed shape within a case highlights the relevant father-child connections. The legend on the right clarifies the color coding and respective roles for nodes C, F, and P.}
\end{figure}

Of all permutations for traversing the tree, $\pi_{BFS}$ is selected in the paper because it reduces the total number of possibilities for $P_{i}^{\pi_{BFS}} = \{0,1,2,3\}$ at least, as shown in Fig.\ref{pic::BFS} and proofed in the supplementary material, minimizing the complexity of the model for predicting the current node's position in the junction tree.

\subsection{Encoder}
\label{sec:encode}

As is mentioned in section \ref{sec:model}, the molecule is represented as sequences of nodes and its position in the junction tree.

Ignited by previous work\cite{r29}, we classify the features of the node into two categories, centrality and spatial, which represent the nature of the node itself and the position of the node in the whole tree respectively. The input $h_i$ to the encoder can be represented as follows:
\begin{equation}
\begin{split}
\label{eq3}
    h_i^{(0)} &= CE_{\epsilon(x_i)} + PE_{\phi(x_i | JT)},
\end{split}
\end{equation}
where $CE_{\epsilon(x_i)}, PE_{\phi(x_i, JT)}\in \mathbb{R}^H$ represents the embedding specified by the centrality and spatial feature of node $x_i$. $\epsilon(x_i): V \to \mathbb{R}$ denotes the centrality feature of the node, where $\phi(x_i| JT) : V \times V \to \mathbb{R} $ denotes the spatial position in the junction tree.

A parallelization strategy is implemented for JTreeformer's encoder, which is structured by GCN and a multi-head self-attention module. Drawing from the conclusions of Min et al. \cite{r31}, this method effectively integrates the Graph Convolutional Network's (GCN) capacity for harvesting local graph details with the Graph Attention Mechanism's ability to accrue global information, thereby enabling a comprehensive grasp across the entire junction tree.

Denote $Attn_{l}$ as $l^{th}$ attention block, we have:
\begin{equation}
\begin{split}
\label{eq4}
Attn_{l} = (\frac{(h_{l-1}W_Q)(h_{l-1}W_K)^T}{\sqrt{d}} + b_{A_{ij}})h_{l-1}W_V,
\end{split}
\end{equation}
where A denotes the adjacency matrix, $b_{A}$ denotes the learnable bias and $(h_{l})$ denotes the encoding of $l^{th}$ block. Additionally, a trainable virtual node called \texttt{[JNode]} embedding \cite{r29} is concatenated along the sequence at the beginning. This virtual node is deemed adjacent to all nodes in the graph, facilitating the extraction of comprehensive graph information. It can be likened to the \texttt{[CLS]} token in BERT \cite{r40}.

Denote $GCN_{l-1}$ as the $l^{th}$ GCN block \cite{r44}, we have:
\begin{equation}
\begin{split}
\label{eq5}
GCN_{l} &= \sigma(\hat{D}^{-1/2}\hat{A}\hat{D}^{-1/2}h_{l-1}W_G), \\
\hat{A} &= A + I, \ \hat{D} = diag(\hat{A}),
\end{split}
\end{equation}
where $W_G$ denotes a learnable weight matrix, $diag(\cdot)$ refers to placing vectors on the diagonal of the matrix, and $\sigma(\cdot)$ denotes the activate function.

Finally, the output result is concatenated with the outcome of multi-head self-attention. Following a layer of linear transformation, it is forwarded to the feed-forward module:
\begin{equation}
\begin{split}
\label{eq6}
\Bar{X}_{l-1} &= GCN_{l-1} \odot Attn_{l-1}, \\
X_l &= \sigma(\Bar{X}_{l-1}W_A)W_B,
\end{split}
\end{equation}
where the matrices $\rm W_{A}$,$\rm W_{B}$ denote trainable weights, and $\odot$ denotes the concatenate process.
We utilize the latent vector of \texttt{[JNode]} as the representation of the molecule, which encapsulates the complete information of the junction tree.

\subsection{Decoder and Loss}
\label{sec:decode}

To gather information in the latent space generated by the encoder, the JTreeformer decoder has a similar structure to the encoder. As discussed in the previous section \ref{sec:model}, the decoder mainly contains position and junction prediction for constructing a junction tree. For the sequence $s_{i+1}^\pi = (J_{i+1},P_{i}^\pi) \in \mathbf{R}^{V_J \times P}$, $J_{i} \in V_J$ and its conditional distribution $p(S^\pi)$, we have:
\begin{equation}
\resizebox{.85\linewidth}{!}{$
\begin{aligned}
\label{eq7}
    p(S^\pi) &=\prod_{i=1}^{n+1} p(s_{i}^\pi | s_1^\pi,\dots,s_{i-1}^\pi) = \prod_{i=1}^{n+1} p_J(J_{i} | J_1,\dots,J_{i-1}) \cdot p_{pos}(P_{i}^\pi | P_1^\pi,\dots,P_{i-1}^\pi),
\end{aligned}
$}
\end{equation}
where $P_{i}^{\pi_{BFS}} = \{0,1,2,3\}$ for the possible position of the predicting node, and $J_{i} \in V_J$ for possible junction in the vocabulary $V_J$. After $k$ nodes are predicted, the $(k+1)^{th}$ junction's position and type will be predicted by the decoder according to the previous junction tree, until $s_{n+1}^\pi$ as the end of sequence token \texttt{[EOS]} is predicted.

The architecture of the embedding mechanism in the decoder is similar to the encoder, and we adapt a masked multi-head attention structure consulting of previous work\cite{r30}. However, GCNs cannot predict the next node without prior knowledge of the entire graph structure, so it's not well adapted to the decoder tasks. To compensate for the problem, we devise a new convolutional network toward graph generation called Directed Acyclic Graph Convolution Network (DAGCN). Denote $DAGCN_{l}$ as the $l^{th}$ DAGCN block of decoder, we have:
\begin{equation}
\begin{split}
\label{eq8}
DAGCN_{l} &= \sigma(Kh_{l-1}W_G), \ \ K = I + \theta D^{-1/2}(D-M)D^{-1/2}, \\
M &= f_M(A),\ \ \ \ D = diag(M),
\end{split}
\end{equation}
where $\theta$ is a learnable parameter, $f_M$ is a mapping utilized to mask the prevent positions from attending to subsequent positions, $A$ denote the adjacency matrix, and $h_{l-1}$ denote the output of the previous layer. The DAGCN's result will be concatenated with Masked Multi-head attention over a softmax layer for the distribution $p_{pos}$ and $p_{J}$. The DAGCN algorithm can be applied to the task of node prediction for arbitrary directed acyclic graphs. DAGCN has a more stable training process than traditional GCN, which is proofed in supplementary material.

To enhance the extraction of molecular properties, an auxiliary prediction task is introduced, aiming to predict relative molecular weight ($W$), relative lipid solubility ($logP$), and topological polarity surface area (TPSA) of molecules using the encoding of virtual nodes. Denote $PRE_{aux}$ as the linear blocks for the prediction task, we have:
\begin{equation}
\begin{split}
\label{eq9}
    C = PRE_{aux}(h_{L}),
\end{split}
\end{equation}
where $h_{L}$ denotes the latent vector of \texttt{[JNode]} representing the junction tree, and $C$ denotes the vector of the representative feature of the junction tree.

The trained loss function encompasses reconstruction loss $\mathbb{E}_q(p)$ and the deviation of auxiliary tasks. 
 Specifically, we do not consider the KL loss compared to traditional VAE. The diffusion model for latent space learning, which is described in the next subsection \ref{sec:diff}, efficiently captures the data distribution without approximating the latent space vectors to the standard normal distribution. The overall loss function can be expressed as:
\begin{equation}
    \resizebox{.85\linewidth}{!}{$
\begin{split}
\label{eq10}
    \mathcal{L}(\theta, \phi; x) &= \mathbb{E}_{q_\phi(z|x)}[\log p_\theta(x|z)] + \delta L_{aux} \\
    \mathbb{E}_{q_\phi(z|x)}[\log p_\theta(x|z)] &= \alpha L_{father} + \beta  L_{curr} = \alpha \sum_{i=1}^n y_{i} \ln \hat{y}_{i,p} + \beta  \sum_{i=1}^n u_{i} \ln \hat{u}_{i,p},
\end{split}
$}
\end{equation}
where the predicted mean squared error (MSE) serves as the loss for auxiliary task $L_{aux} = \sum_{i=1}^n(C-\hat{C})^2$. For the reconstruction loss $\mathbb{E}_q(p)$, we consider the probability of the father's node $y$ and the current junction $u$. Let $\phi, \theta$ denote the encoder's and decoder's parameters. $y_{i}/u_{i}$ denotes the original distribution of the father/current of $i^{th}$ node, and $\hat{y}_{i,p}/\hat{u}_{i,p}$ denotes the decoder's distribution of the father/current of $i^{th}$ node. $q_\phi(z|x)$ denotes the conditional distribution of latent space under the original data $x$, and $p_\theta(x|z)$ denotes the decoder's distribution under the latent vector $z$. We take $\rm \alpha$=$\rm \beta$=1, $\rm \delta$=0.2.

\subsection{Skip-connected Latent Diffusion}
\label{sec:diff}



With the trained encoder and decoder, we further insert a DDIM\cite{r24} for latent space learning, which is a deterministic diffusion process that enables efficient and controlled generation of molecular structures, by iteratively refining them from a noise distribution towards a predictable data distribution.

\begin{algorithm}[t]
	\caption{A step of pre-train process in JTreeformer}
        \label{alg::train}

    \begin{algorithmic}[1]
    \REQUIRE{Original molecule distribution $\phi$, number of model layers $L$, JTreeformer decoder $\mathbb{D}$, lower boundary of loss $\epsilon$}
	\ENSURE{Pretrained JTreeformer}
    \FOR{each molecule graph $G_i(V,E) \in \phi$}
        \STATE Encoding sequence $S_i^\pi \leftarrow f_{J\rightarrow S}(f_{G\rightarrow J}(G_i)|\pi)$ (Eq.\eqref{eq1}, Eq.\eqref{eq2})
        \STATE Encoding vector for each node $h_i^{(0)} \leftarrow CE_{\epsilon(x_i)} + PE_{\phi(x_i | JT)}$ (Eq.\eqref{eq3})
        \FOR{each layer $l < L$}
            \STATE Calculate $Attn_l, GCN_l$ (Eq.\eqref{eq4}, Eq.\eqref{eq5})
            \STATE Calculate $h_l \leftarrow GCN_{l-1} \odot Attn_{l-1}$ (Eq.\eqref{eq6})
        \ENDFOR
        \STATE $X_L^{(i)} \leftarrow \sigma(h_LW_A)W_B \in \Gamma$  (Eq.\eqref{eq6})
    \ENDFOR
    \STATE Define latent space representation $\Gamma$ where $X_L \in \Gamma$ 

    \WHILE{$\hat{s}_{n+1}^\pi$ is not \texttt{[EOS]}}
        \STATE Next node $\hat{s}_{n+1}^\pi \leftarrow \mathbb{D}(X_L | (s_1^\pi,\dots, s_n^\pi))$
    \ENDWHILE
    \STATE Generated sequence $\hat{S}^\pi = (\hat{s}_{0}^\pi,\dots,\hat{s}_{n+1}^\pi)$
    \STATE Calculate $f_{loss}(\hat{S}^\pi,S^\pi)$  (Eq.\eqref{eq10})
     \IF{$f_{loss} < \epsilon$} 
     \STATE Train Diffusion model in latent space $\Gamma$ (Eq.\eqref{eq12}, \eqref{eq13})
     \ENDIF
    \end{algorithmic}
\end{algorithm}

Given the $t$ step of diffused latent vector $h_t$, we can predict $h_{t-1}$ as:
\begin{equation}
    \resizebox{.92\linewidth}{!}{$
\begin{split}
\label{eq11}
h_{t-1} &= \sqrt{\frac{\alpha_{t-1}}{\alpha_t}} \left( h_t - \frac{1-\alpha_t}{\sqrt{\alpha_t}} \epsilon_\theta(h_t, t) \right) + \sqrt{1 - \alpha_{t-1} - \sigma_t^2} \cdot \epsilon_\theta(h_t, t) + \sigma_t \epsilon_t
\end{split}
$}
\end{equation}

\begin{equation}
\label{eq12}
\sigma_t^2 = \eta\cdot\widetilde{\beta}_t = \eta \cdot \sqrt{(1-\alpha_{t-1})/(1-\alpha_t)}\sqrt{(1-\alpha_t/\alpha_{t-1})},
\end{equation}
where $\alpha_t, \alpha_{t-1}$ denote the coefficient defined by step $t$, $\epsilon_t$ represents the noise unrelated to $h_t$.  

We employ a skip-connected NLP\cite{r-1} as the noise prediction network $\epsilon_\theta(\cdot)$ instead of U-net. The traditional U-net is constructed by mainly two parts, the down-sampling and the up-sampling part, consisting of several convolution layers. Denote $DW$ as the down-sampling convolution layer and $UP_{SC}$ as the up-sampling layer with skip-connected structure, we have:
\begin{equation}
\label{eq13}
\begin{split}
d_i = DW(d_{i-1}), u_i = UP_{SC}(u_{i-1}, d_{i-1}),i=1,\dots,N.
\end{split}
\end{equation}

With skip-connected mechanism\cite{r-2}, the network can aggregate the information of each down-sampling layer to corresponding up-sampling layers, which has more affirmative efficiency of noise prediction. 

\subsection{Molecular Generation}

After sampling from the latent space using the Diffusion Model, the JTreeformer is employed to reconstruct the junction tree. The assembly of the junction tree into molecules begins at the root and progressively attaches fragments represented by the children of the current node to the previously integrated molecules, following a DFS order. This process continues until all nodes are assembled or until some nodes cannot be attached to the existing molecules. The attachment of fragments to molecules can result in multiple isomers, necessitating a selection process. While JT-VAE relies on a deep neural network for isomer selection, which can be computationally expensive, we propose using Monte Carlo Tree Search (MCTS) as an alternative. MCTS treats isomer selection as a decision problem and, through Monte Carlo simulation and state space pruning, ensures that the final assembled molecule achieves the highest score. In our experiments, scoring is determined by two factors: 1) likelihood scoring, where a higher number of assembled nodes leads to a higher score; and 2) properties scoring, where the assembled molecular properties (e.g., logP) are evaluated based on their proximity to the target values.

\begin{algorithm}[t]
	\caption{Decoding in JTreeformer}
        \label{alg::test}
     \begin{algorithmic}[1]
     	\REQUIRE{Normal distribution $\mathbb{N}(0,I)$, pretrained latent diffusion model $\mathbb{L}$ and decoder $\mathbb{D}$, MCTS model $M$}
	\ENSURE{Generated molecule distribution $\mu$}
        \STATE $\alpha \in \mathbb{N}(0,I)$
        \STATE Latent representation $L = \mathbb{L}(\alpha)$
        \WHILE{$\hat{s}_{n+1}^\pi$ is not \texttt{[EOS]}}
            \STATE Next node $\hat{s}_{n+1}^\pi \leftarrow \mathbb{D}(L | (s_1^\pi,\dots, s_n^\pi))$
        \ENDWHILE
        \STATE Generated sequence $\hat{S}^\pi = (\hat{s}_{0}^\pi,\dots,\hat{s}_{n+1}^\pi)$ 
        \STATE Molecule distribution $\mu = M(\hat{S}^\pi)$
         
     \end{algorithmic}
\end{algorithm}

\section{Experiments}

Our evaluation efforts contain JTreeformer's performance, validation, and ablation section. 
Experiments across three distinct segments are conducted, with a variety of metrics utilized to assess the comparative performance of the models.

\begin{itemize}
  \item \textbf{Performance for molecule generation.} The JTreeformer was utilized in comparative analyses with several baseline models, employing multiple evaluation criteria to determine the generation quality.
  \item \textbf{Validation for latent space and diffusion.} Visualization efforts have been applied to demonstrate the generated latent space, the diffusion sampling process, and the neighbors of specific molecules produced by JTreeformer.
  \item \textbf{Ablation study.} We conduct ablation studies by selectively removing certain components of the model to investigate their respective contributions to its overall functionality.
\end{itemize}

Below we describe the data, baselines, and metrics of all tasks included. 

\textbf{Data:} We trained our model on the MOSES\cite{r51} and QM9\cite{qm9} dataset, encompassing 1.6 million filtered Lead Like molecules derived from the ZINC dataset, providing a benchmark for comparison.

\textbf{Baselines:} For the MOSES dataset, we compare our approach with a range of established baselines including: 1) Hidden Markov Models (HMM)\cite{r51}, 2) NGram models\cite{r51}, 3) Combinatorial \\ generator\cite{r51}, 4) CharRNN\cite{r53,r54}, 5) Variational Autoencoders (VAE)\cite{r55}, 6) Adversarial Autoencoders (AAE)\cite{r56}, 7) Junction Tree Variational Autoencoder (JTN-VAE)\cite{r8}, and 8) LatentGAN\cite{r57}. For the QM9 dataset, we compare our approach with baselines provided in \cite{final}.

\textbf{Metrics:} We employ Valid, Unique, Novelty, and Internal diversity ($IntDiv_p$), to evaluate generation quality, where the first three denote the valid, non-repetitive, and new molecule generated by the model and Internal diversity is defined as:
\begin{equation}
IntDiv_p(G) = 1 - \sqrt[p]{\frac{1}{|G|^2}\sum_{m_1,m_2 \in G}T(m_1,m_2)^p},
\end{equation}
where T($\cdot$,$\cdot$) represents the Tanimoto similarity between the fingerprints of two molecules. A lower IntDiv indicates poorer diversity within the generated molecules and the collapse of latent space.

\textbf{Training details:} We performed junction tree decomposition on over 1.7 million molecules from the MOSES dataset, where 1 NVIDIA A100 is utilized during training process, resulting in a vocabulary $\mathbb{V}$ of size $|\mathbb{V}| = 496$. The JTreeformer used in the main text had 12 layers for both the encoder and decoder. The encoder had a hidden dimension of 512, and the feed-forward layers in the JTreeformer blocks had an expansion dimension of 1024. The decoder had a hidden dimension of 768, and the feed-forward layers in the JTreeformer blocks had an expansion dimension of 1536. The maximum number of hydrogen number allowed as input was set to 50, the maximum node degree was set to 20, and the maximum number of layers was set to 50.  
  
The JTreeformer was trained on MOSES for 4 epochs with a learning rate of 0.0001, a batch size of 256, and a warm-up strategy where the learning rate linearly increased from 0 to 0.0001 during the first epoch and then decayed exponentially at a rate of 3.036e-4 at each step. The parameters of the JTreeformer trained with different KL loss weights in the supplementary materials were the same as those in the main text, but were trained for 8 epochs on MOSES with a batch size of 256. During the first 4 epochs, the KL loss weight was set to 0, and then linearly increased to its maximum value during the next 4 epochs. Training also employed a warm-up strategy with the same learning rate schedule as mentioned above.  
  
After mapping the MOSES dataset to the latent space using the pre-trained JTreeformer, we trained DDIM in the latent space. DDIM used a skip connection network with 6 layers, a hidden dimension of 768, and a maximum diffusion time step of 1000. The loss function was the Mean Absolute Error (MAE), and the model was trained for 20 epochs with a batch size of 256 and a learning rate of 0.0001. Training employed a warm-up strategy with the learning rate linearly increasing from 0 to 0.0001 during the first epoch and then decaying exponentially at a rate of 5.593e-5 at each step.  

\subsection{Performance for molecule generation}

Table \ref{table::1} shows that our MOSES set JTreeformer outperforms previous modela in molecule generation quality on MOSES se. Notably, JTreeformer significantly outperforms the previous on IntDiv metric, indicating its strong ability in molecular structures learning. This enhancement likely stems from the model's comprehensive utilization of both local and global graph information. Additionally, our model surpasses most others on Novelty, demonstrating its ability to discern between sampling and training data. This discriminatory power is attributed to the semantically rich feature construction enabled by the transformer architecture within our model.
\setlength{\tabcolsep}{3pt}
\begin{table}[t]
\footnotesize
\centering
\caption{Performance for baselines and JTreeformer on Moses dataset. Reported mean over three independent model initializations. }
\label{table::1}
\begin{tabular}{lccccc}
\toprule
Model & Valid ($\uparrow$) & Unique ($\uparrow$) & Novelty ($\uparrow$) & $\rm IntDiv_1$($\uparrow$) & $\rm IntDiv_2$($\uparrow$)\\
\midrule
HMM\cite{r51}         & 0.076 & 0.623  & 0.9994 & 0.8466 & 0.8104\\
NGram\cite{r51}       & 0.2376 & 0.974  & 0.9694 & 0.8738 & 0.8644\\
Combinatorial\cite{r51} & \textbf{1.0} & 0.9983 & 0.9878 & 0.8732 & 0.8666\\
CharRNN\cite{r53,r54}     & 0.975 &  \textbf{1.0} & 0.8942 & 0.8562 & 0.8503\\
VAE\cite{r55}         & 0.977 & \textbf{1.0} & 0.6895 & 0.8557 & 0.8499\\
AAE\cite{r56}         & 0.937 & \textbf{1.0} & 0.7993  & 0.8558  & 0.8498\\
JTN-VAE\cite{r8}     & \textbf{1.0} & \textbf{1.0}  & 0.9143  & 0.8561  & 0.8493 \\
LatentGAN\cite{r57}   & 0.897 & \textbf{1.0} & 0.949  & 0.8565  & 0.8505\\
\midrule
JTreeformer & \textbf{1.0} & 0.986 & 0.9988 & \textbf{0.8822} & \textbf{0.8724}\\
\bottomrule
\end{tabular}
\end{table}

\setlength{\tabcolsep}{3pt}
\begin{table}[t]
\footnotesize
\centering
\caption{Performance for baselines (data from \cite{final}) and JTreeformer on QM9 dataset. Uniq refers to Uniqueness.}
\label{table::k}
\begin{tabular}{lcccccc}
\toprule
Model & Uniq ($\uparrow$) & Novelty ($\uparrow$) & KL Div ($\uparrow$) & FCD ($\uparrow$) \\
\midrule
JT-VAE    & 0.549 & 0.386  & 0.891  & 0.588  \\
GCN       & 0.533 & 0.320  & 0.552  & 0.174  \\
GraphAF   & 0.500 & 0.453  & 0.761  & 0.236  \\
GraphDF   & 0.672 & 0.672  & 0.601  & 0.137  \\
GA        & 0.008 & 0.008  & 0.649  & 0.540  \\
MARS      & 0.659 & 0.612  & 0.547  & 0.123  \\
HierVAE   & 0.416 & 0.285  & 0.719  & 0.358  \\
H-VAE*    & 0.619 & 0.487  & 0.869  & 0.588  \\
F-VAE*    & 0.466 & 0.400  & 0.797  & 0.443  \\
PS-VAE    & 0.673 & 0.523  & \textbf{0.921}  & \textbf{0.659}  \\
\midrule
JTreeformer (ours) & \textbf{0.785} & \textbf{0.758} & 0.910 & 0.312 \\
\bottomrule
\end{tabular}
\end{table}

\begin{figure*}[h!]
  \centering
  \includegraphics[width=\linewidth]{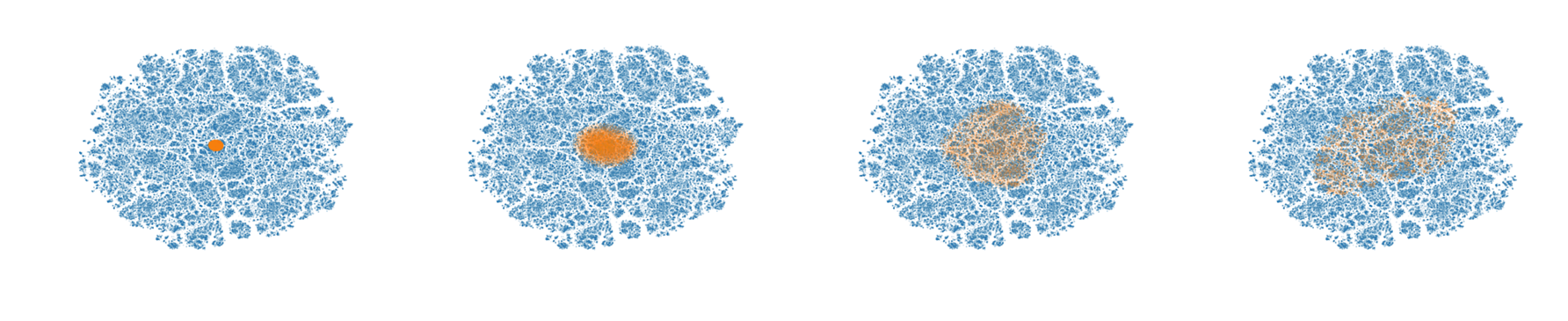}
  \caption{Sampling by pre trained DDIM model in Latent space generated by JTreeformer in different diffusion steps of 50, 100, 200, 1000 of removing noise from the original samplings.}
  \Description{The image showcases a sequence of four circular patterns, each containing numerous small dots. The dots are primarily blue in color, with a few orange dots scattered among them. The patterns appear to be generated by a computational model, specifically a pre-trained DDIM model in latent space, as indicated by the accompanying text. The purpose of the patterns seems to be related to sampling and noise removal from original data.}
  \label{fig::latent space}
\end{figure*}


\subsection{Validation for latent space and diffusion}
We visualize the sampling process of the Diffusion Model, observe patterns in the latent space, and perform semantic interpolation in the latent space.

\textbf{Sampling steps. }Fig.\ref{fig::latent space} showes that during the denoising process, the sampling points progressively extend into the distribution domain of the MOSES and QM9 set, circumventing the out-of-distribution realm, thereby attesting to DM's proficiency in learning the distribution of latent vectors.

\begin{figure}[h!]
  \centering
  \includegraphics[width=\linewidth]{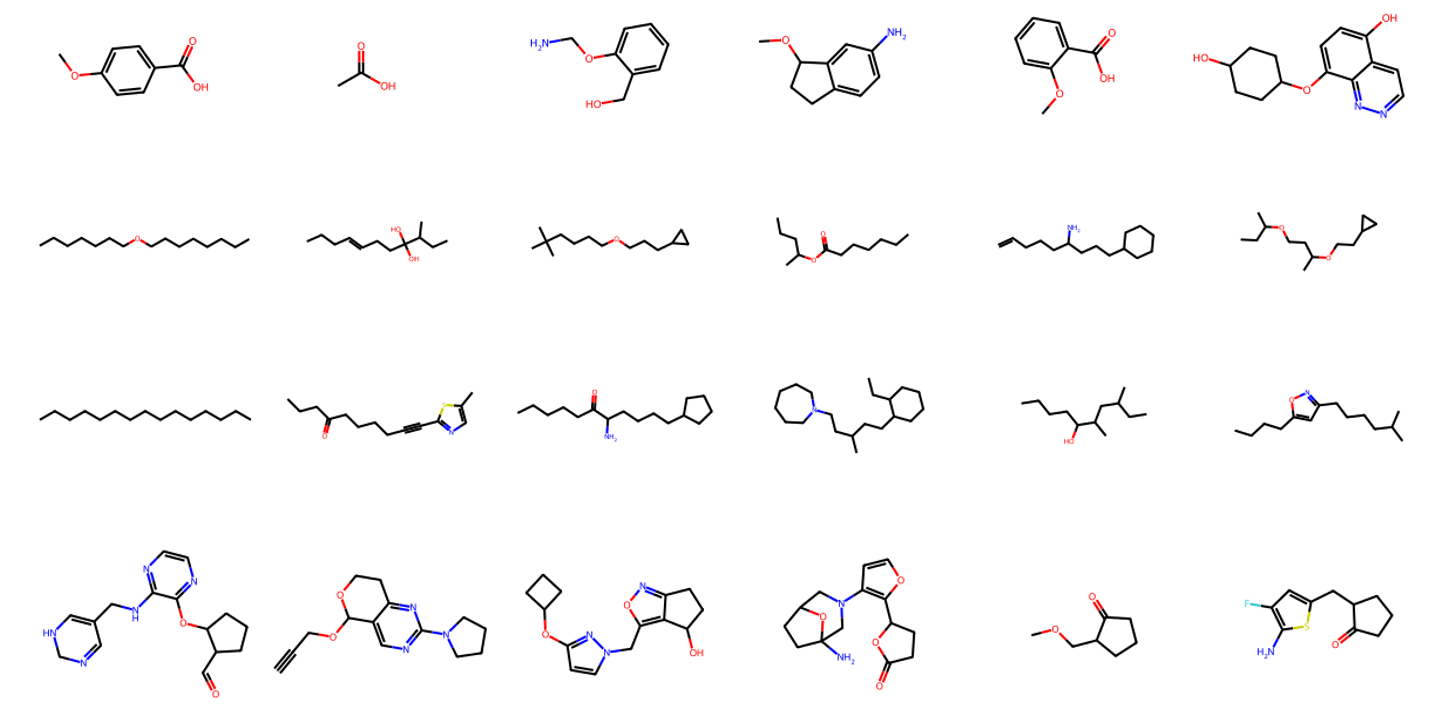}
  \caption{The leftmost one is the input molecule, followed by their potential spatial neighbors, which can be seen to have similar scaffolding and functional groups to the input molecule.}
  \label{fig::noisedsample}
  \Description{The image displays a series of molecular structures represented in a grid format. Each molecule is depicted using lines to indicate bonds between atoms, with different colors denoting various types of atoms or functional groups. A text label at the bottom explains that the last molecule on the left is the input molecule, and those following it are potential spatial neighbors, sharing similar scaffolding and functional groups with the input molecule.}
\end{figure}

\begin{figure*}[h!]
  \centering
  \includegraphics[width=\linewidth]{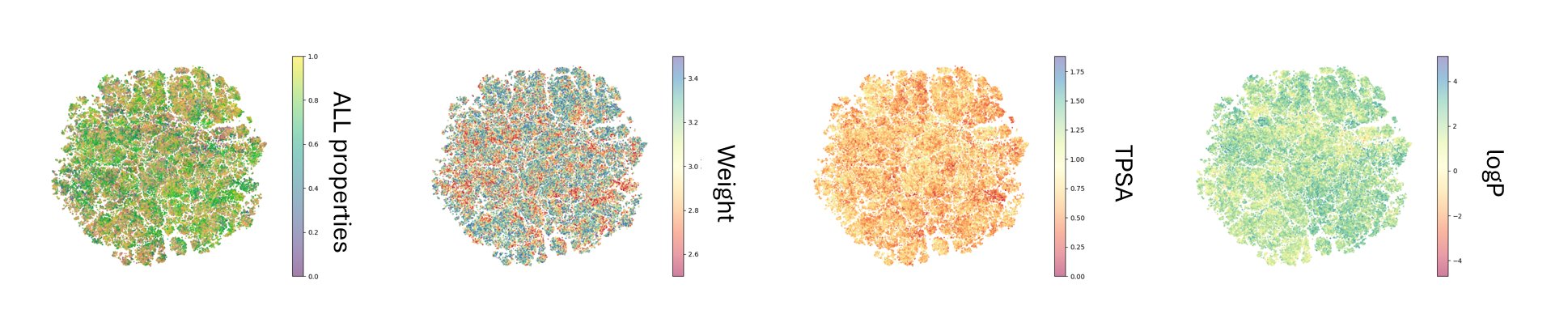}
  \caption{Latent space generated by JTreeformer. Different colour is utilized to represent properties including the Weight, TPSA, logP of molecules. A synthesized figure is displayed first, which takes the value of Weight, TPSA, logP as RGB separately.}
  \label{fig::latentspace2}
  \Description{The image displays the latent space generated by JTreeformer, showcasing molecular properties. Each of the five panels represents a different visualization:
This first plot displays a comprehensive view of the latent space, using a color gradient to represent the combination of multiple molecular properties. The color scale ranges from low to high values, with a gradient indicating property distribution across the space.
 The second panel highlights the weight property of molecules, where the color scale reflects varying molecular weights. The gradient suggests a correlation between the latent space structure and molecular mass. The third plot represents the Total Polar Surface Area (TPSA) property of the molecules. Similar to the weight visualization, it uses a color gradient to distinguish varying TPSA values.
The fourth panel depicts the logP values, which represent the octanol-water partition coefficient, again shown with a gradient indicating the variation across the latent space.
 This synthesized figure combines the properties of Weight, TPSA, and logP using RGB colors to encode each of these properties. The red, green, and blue components represent the respective properties in a manner that visually integrates them into a single representation.

Overall, the image emphasizes how the molecular properties are spatially distributed in the latent space and how their values are encoded through color gradients.}
\end{figure*}

\textbf{Latent Patterns. }Empirical evidence suggests that high-performing VAEs encode analogous molecules into a distinct latent space region, forming patterns. To investigate JTreeformer's ability to learn these patterns, we encode molecules, introduce noise to their latent representations, and subsequently decode them. As illustrated in Fig.\ref{fig::noisedsample}, the decoded molecules frequently exhibit scaffolds or functional groups resembling the originals, suggesting that neighbors within the latent space exhibit similar patterns, which are depicted as distinct clusters in Fig.\ref{fig::latentspace2}, clearly delineating the molecular properties via clustered patterns.

\begin{figure}[h!]
  \centering
  \includegraphics[width=\linewidth]{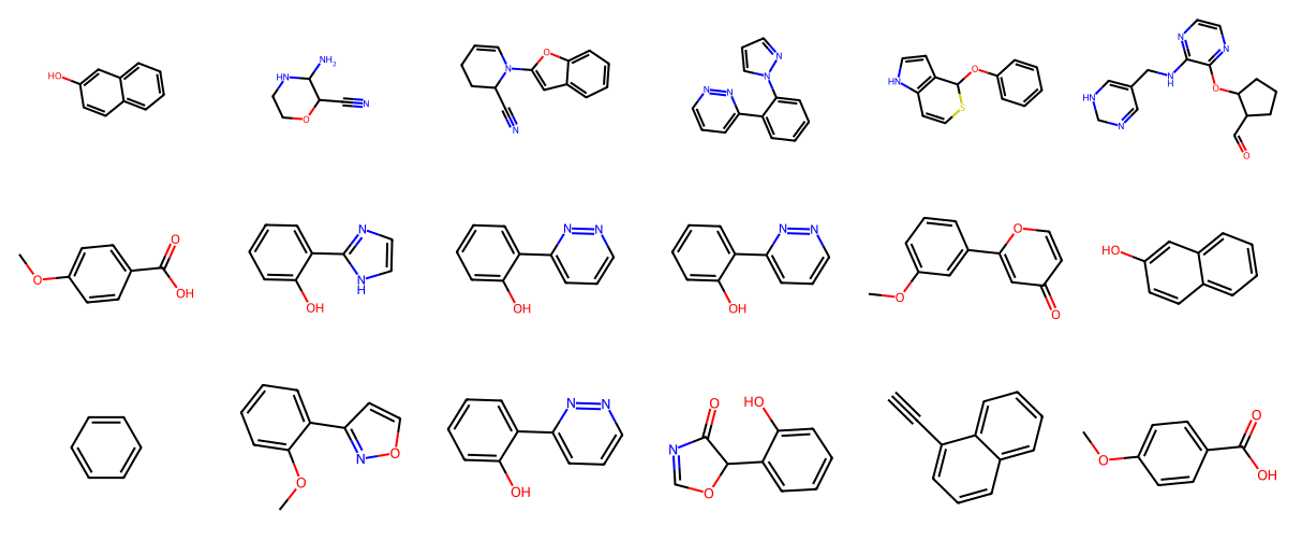}
  \caption{The first and last columns are the starting and ending molecules for chemical molecule interpolation, and the middle ones are the interpolated molecules by steps.}
  \label{fig::si}
  \Description{The image shows the process of chemical molecule interpolation. The first and last columns display the starting and ending molecules, while the middle columns show the interpolated molecules at each step. The molecules evolve smoothly from the initial structure to the final one, with intermediate structures captured in each step of the transformation. The colors in the molecular structures highlight specific atoms or functional groups involved in the interpolation process.}
\vspace{-13pt}
\end{figure}

\textbf{Interpolation for molecules. }In the latent space generated by the JTreeformer, continuous points exhibit continuity of chemical structure, facilitating the integration of diverse inputs and enabling chemical interpolation. To assess JTreeformer's capacity for interpolation of molecules, we map distinct pairs of molecules to latent spaces, select four instances in between two molecules, and weigh the latent vectors of the molecules before decoding. Fig.\ref{fig::si} illustrates interpolation typically yields a composite of the two molecules or reflects a compromise between their scaffolds and functional groups, showing JTreeformer's proficiency in interpolation of molecules.
\begin{figure}[h!]
  \centering
  \vspace{-1em}
  \begin{minipage}[b]{0.9\linewidth}
    \centering
    \includegraphics[width=\linewidth]{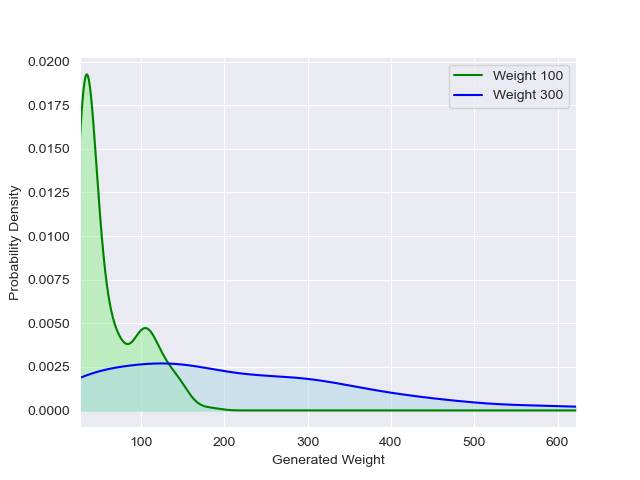}
  \end{minipage}
 \caption{Property optimization with relative molecular mass}
 \Description{This image shows a property optimization graph with respect to relative molecular mass. The x-axis represents the generated weight, while the y-axis shows the probability density. The graph displays two lines, one for a weight of 100 and another for a weight of 300. The graph illustrates how the property optimization changes as the relative molecular mass is varied.}
  \label{fig:image1}
\end{figure}

\textbf{Addtional experiments for property optimization.} We are adding property information by introducing cross-attention in the decoder, and also trying guidance in the latent diffusion.We use relative molecular mass 100 and 300 as the input property to generate the molecule as Fig. \ref{fig:image1}.

\subsection{Ablation study}
We perform a series of ablation studies on the importance of designs in our proposed JTreeformer on MOSES dataset. \\

\textbf{Diffusion Model sampling in latent space.} We evaluate the combined performance of the JTreeformer and Diffusion Model (DM) sampling, and employ our model and the default VAE from MOSES. Table \ref{table::3} demonstrates that MOSES' VAE does not significantly improve performance through DM sampling, whereas  \\JTreeformer exhibits notable enhancement. Moreover, JTreeformer's generation quality significantly improved with DM, highlighting the benefits of the combination of both methods.\\

\setlength{\tabcolsep}{3pt}
\begin{table}[t]
  \centering
  \caption{Ablation study for Diffusion Model sampling.}
  \label{table::3}
  \begin{tabular}{cc|ccccc}
    \toprule
  Diffusion & Jtreeformer & Valid & Unique & Novelty & $\rm IntDiv_1$ & $\rm IntDiv_2$ \\

    \midrule
    \checkmark & \checkmark &\textbf{1.0}&0.986&0.9988&\textbf{0.8822}&\textbf{0.8725}\\
    - & \checkmark &\textbf{1.0}&0.976&\textbf{0.9999}&0.8518&0.8452\\
    \checkmark & - &0.9717&\textbf{1.0}&0.7004&0.8554&0.8485\\
    - & - &0.974&\textbf{1.0}&0.6985&0.8565&0.8496\\
    \bottomrule
  \end{tabular}
\end{table}
\textbf{JTreeformer modules.} We do some ablation study towards DAGCN and features selected from the junction tree (e.g. degree).  Table \ref{table::4} reveals that while Graph Transformer contributes significantly to JTreeformer's capabilities, our proposed feature selection also gathers more information from the tree, which leads to a superior result. Without DAGCN, JTreeformer's performance decreases significantly, inferior even to a model devoid of both feature composition and DAGCN. This indicates that DAGCN enhances feature learning robustness, and more feature selection ensures the model's better training performance when DAGCN is utilized.

\subsection{Molecule Interpolation}
We map distinct pairs of molecules into latent spaces and subsequently select four instances situated between the two molecules. Prior to decoding, we assign weights to the latent vectors of these molecules. Fig. \ref{fig::si} presents additional results of demonstrating molecular interpolations.
These interpolations can be categorized into two types: mixtures of scaffolds and functional groups derived from the pair of molecules, or the appearance of specific chemical fragments originating from the pair of molecules. The former displays the fusion of distinct patterns, while the latter indicates that the patterns encode fragmentary information of molecules, which is retrieved during the interpolation process.


\setlength{\tabcolsep}{3pt}
\begin{table}[t]
  \centering
  \caption{Ablation study for JTreeformer modules.}
  \label{table::4}
  \begin{tabular}{cc|ccccc}
    \toprule
  Feature & DAGCN & Valid & Unique & Novelty & $\rm IntDiv_1$ &$\rm IntDiv_2$ \\

    \midrule
    \checkmark & \checkmark &\textbf{1.0}&\textbf{0.986}&0.9988&\textbf{0.8822}&\textbf{0.8725}\\
    - & \checkmark &\textbf{1.0}&0.963&0.9983&0.8782&0.8489\\
    \checkmark & - &\textbf{1.0}&0.176&\textbf{1.0}&0.6059&0.5615\\
    - & - &\textbf{1.0}&0.918&0.9978&0.8795&0.8586\\
    \bottomrule
  \end{tabular}
  \vspace{-13pt}
\end{table}

\section{Conclusion}

This paper introduces JTreeformer, a pioneering graph-transformer-based framework for molecular generation. JTreeformer uniquely combines paralleled Graph Convolutional Networks (GCN) with multi-head self-attention mechanisms to effectively capture both global and local molecular information. This dual approach allows the model to learn a semantically rich latent space, where intricate relationships between atoms, bonds, and substructures are encoded. This latent space representation enhances the model's ability to generate novel and diverse molecular structures that are both chemically valid and biologically relevant. A notable feature of JTreeformer is its use of Directed Acyclic Graph Convolutional Networks (DAGCN) integrated with masked multi-head attention in the decoder. This architecture enables the model to iteratively synthesize the complete molecular structure, preserving chemical integrity while improving generation quality.
Furthermore, JTreeformer introduces the innovative use of a diffusion model (DM) trained in the latent space of molecular generation. By capturing the complex distribution of chemical space, the diffusion model significantly improves both the efficiency and effectiveness of molecular generation, allowing for better exploration of chemical space. The experimental results demonstrate the superiority of JTreeformer, providing new insights and strategies for molecular discovery, with potential applications in drug design and materials science.

\bibliographystyle{ACM-Reference-Format}
\bibliography{icmr}

\end{document}